\pdfoutput=1
\documentclass[11pt]{article}
\usepackage[preprint]{acl}
\usepackage{times}
\usepackage{array}
\usepackage{latexsym}
\usepackage{amsmath}
\usepackage{amssymb}
\usepackage[T1]{fontenc}
\usepackage[utf8]{inputenc}
\usepackage{microtype}
\usepackage{inconsolata}
\usepackage{graphicx}
\usepackage{pifont}
\usepackage[most]{tcolorbox}
\usepackage{times}
\usepackage{latexsym}

\usepackage{amsthm}
\usepackage[T1]{fontenc}
\usepackage{bbm}
\usepackage{pifont}
\usepackage[utf8]{inputenc}
\usepackage{microtype}
\usepackage{makecell}
\usepackage{multirow}
\usepackage{setspace}
\usepackage{booktabs}
\usepackage{threeparttable}
\usepackage{CJKutf8}
\usepackage{enumerate}
\usepackage{amsfonts,amssymb,amsmath} 
\usepackage{color}
\usepackage{graphicx}
\usepackage{tabularx} 
\usepackage{siunitx}
\usepackage{caption}
\usepackage{algorithm}
\usepackage{algorithmic}
\usepackage{multirow}
\usepackage{hyperref}
\usepackage{url}
\usepackage{booktabs} 
\usepackage{wrapfig}
\usepackage{algorithm}
\usepackage{caption}
\usepackage{subcaption}
\usepackage{amsmath}
\usepackage{amssymb}
\usepackage{mathtools}
\usepackage{cases}
\usepackage{mathrsfs}
\usepackage{subeqnarray}
\usepackage{multirow}
\usepackage{newfloat}
\usepackage{listings}
\usepackage{enumitem}
\usepackage{bbm}
\usepackage{etex}
\usepackage{xspace}
\usepackage[etex=true,export]{adjustbox}
\usepackage{authblk}
\usepackage[capitalize,noabbrev]{cleveref}
\usepackage{graphicx}
\usepackage{arydshln}
\usepackage{array,calc}
\definecolor{myyellow}{RGB}{255,255,204}
\definecolor{mygrey}{RGB}{128,128,128} 
\definecolor{myblue}{RGB}{221, 233, 247}
\newcommand{\classbg}[1]{\setlength{\fboxsep}{1pt}\colorbox{myblue}{\textbf{#1}}}

\lstdefinelanguage{mycase}{
    basicstyle=\scriptsize\ttfamily, 
    moredelim = [s][\color{mygrey}]{\{}{\}},
}

\newtcblisting{showcase}[1][]{
  enhanced,
  arc=0em,
  boxrule=.5pt,
  listing only,
  listing options={
    language=mycase,
    upquote=true,
    basicstyle=\scriptsize \ttfamily \setlength{\baselineskip}{1.1\baselineskip},
    breaklines=true,  
    breakindent=0pt,
    xleftmargin=0pt,
    xrightmargin=0pt,
    aboveskip=-4pt,
    belowskip=-4pt,
    columns=fullflexible,
    escapeinside={|}{|},
  },
  colback=white,
  colframe=gray,
  breakable, 
  colbacktitle=gray!10,        
  coltitle=black,             
  attach boxed title to top center={yshift=-3mm},
  #1
}

\newtcolorbox[list inside=prompt,auto counter,number within=section]{prompt}[1][]{
    colbacktitle=black!60,
    coltitle=white,
    fontupper=\footnotesize,
    boxsep=5pt,
    left=0pt,
    right=0pt,
    top=0pt,
    bottom=0pt,
    boxrule=1pt,
    #1,
}

\title{AGENT-X: Adaptive Guideline-based Expert Network for Threshold-free AI-generated teXt detection}

\author{
 \textbf{Jiatao Li\textsuperscript{1,2}},
 \textbf{Mao Ye\textsuperscript{1}},
 \textbf{Cheng Peng\textsuperscript{1}},
 \textbf{Xunjian Yin\textsuperscript{1}},
 \textbf{Xiaojun Wan\textsuperscript{1}}
\\
 \textsuperscript{1}Wangxuan Institute of Computer Technology, Peking University
\\
 \textsuperscript{2}Information Management Department, Peking University
 \\
 \texttt{{leejames@stu.pku.edu.cn}}, 
 \texttt{wanxiaojun@pku.edu.cn}
}

\begin{document}
\maketitle
\begin{abstract}
Existing AI-generated text detection methods heavily depend on large annotated datasets and external threshold tuning, restricting interpretability, adaptability, and zero-shot effectiveness. To address these limitations, we propose \textit{AGENT-X}, a zero-shot multi-agent framework informed by classical rhetoric and systemic functional linguistics. Specifically, we organize detection guidelines into semantic, stylistic, and structural dimensions, each independently evaluated by specialized linguistic agents that provide explicit reasoning and robust calibrated confidence via semantic steering. A meta agent integrates these assessments through confidence-aware aggregation, enabling threshold-free, interpretable classification. Additionally, an adaptive Mixture-of-Agent router dynamically selects guidelines based on inferred textual characteristics. Experiments on diverse datasets demonstrate that \textit{AGENT-X} substantially surpasses state-of-the-art supervised and zero-shot approaches in accuracy, interpretability, and generalization.
\end{abstract}

\section{Introduction}
The rapid proliferation of texts generated by Large Language Models (LLMs) has intensified the challenge of distinguishing AI-generated content from authentic human writing~\cite{ippolito2020automatic, mitchell2023detectgpt, su2023detectllm}. Traditional detection approaches rely heavily on supervised classifiers trained with extensive labeled datasets and external threshold optimization~\cite{DBLP:conf/coling/JawaharASL20, bakhtin2019real, yan2023detection}, severely limiting their interpretability, adaptability, and effectiveness in realistic zero-shot scenarios, where domain-specific calibration datasets are frequently unavailable~\cite{bao2024fastdetectgptefficientzeroshotdetection, verma2024ghostbuster}.

To address these fundamental limitations, we propose \textit{AGENT-X}, a novel zero-shot multi-agent detection framework rooted in theories drawn from classical rhetoric and systemic functional linguistics~\cite{halliday1978language, simpson2004stylistics}. We first curate an extensive set of detection guidelines derived from academic literature and online communities, explicitly organizing them into semantic, stylistic, and structural linguistic dimensions. This structured theoretical foundation provides a clear, interpretable basis for distinguishing human-written texts from AI-generated content.

\begin{figure*}
\centering
\includegraphics[width=0.95\linewidth]{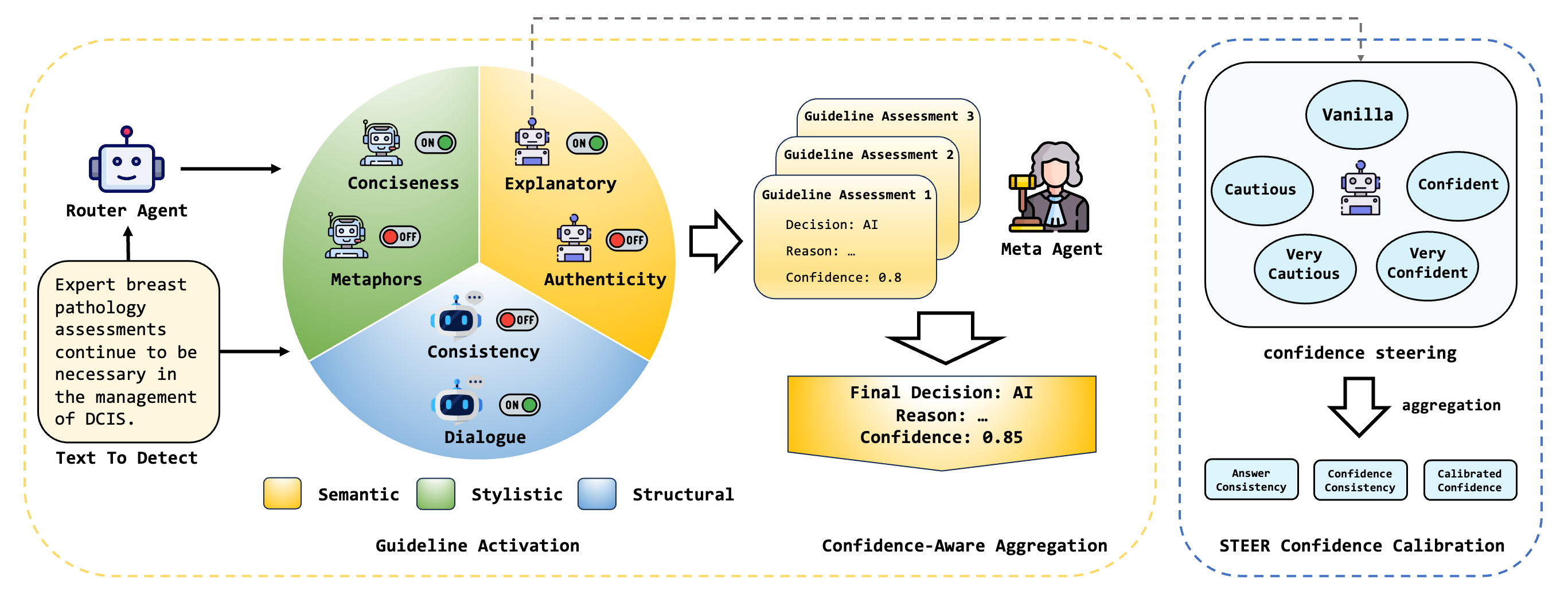}
\caption{Overview of AGENT-X. The router agent dynamically selects relevant linguistic agents, each generating robust, calibrated confidence scores via Confidence Steering. The meta agent aggregates these assessments to classify texts as human-written or AI-generated.}
\label{fig:workflow}
\end{figure*}

As illustrated in Figure~\ref{fig:workflow}, \textit{AGENT-X} employs specialized LLM agents, each independently analyzing texts within their respective linguistic dimensions. These agents generate explicit classification decisions, detailed linguistic reasoning, and robustly calibrated confidence scores via semantic steering methods. Crucially, our framework introduces a meta agent that integrates these dimension-specific evaluations through a confidence-aware aggregation mechanism, eliminating the need for external threshold optimization and significantly enhancing interpretability and practical applicability.

To effectively handle the diversity and complexity of real-world texts, we further incorporate a dynamic Mixture-of-Agent routing mechanism. This adaptive approach selects the most relevant linguistic guidelines based on inferred textual characteristics—such as topical domains and typical textual features—enabling \textit{AGENT-X} to robustly address varied zero-shot detection scenarios.

Moreover, we observe that although existing detection methods commonly report threshold-independent metrics such as AUROC, their practical deployment inherently necessitates selecting thresholds to translate predicted probabilities into actionable binary decisions. This external threshold calibration introduces variability and often leads to inconsistent and suboptimal accuracy performance in real-world applications. In contrast, our proposed multi-agent framework inherently circumvents these calibration challenges through its confidence-aware, threshold-free aggregation mechanism. Consequently, \textit{AGENT-X} substantially improves accuracy, interpretability, and real-world applicability.

Our contributions are as follows\footnote{Our code and curated guidelines will be released to facilitate future research.}:
\begin{itemize}[leftmargin=*,itemsep=1pt]
    \item We introduce the first comprehensive, theoretically-grounded categorization of AI-generated text detection guidelines into explicit linguistic dimensions, enabling interpretable stylistic analysis.
    \item We propose \textit{AGENT-X}, a zero-shot multi-agent framework integrating adaptive guideline routing, explicit linguistic reasoning, and advanced semantic-steering confidence calibration, significantly outperforming existing supervised and zero-shot methods.
    \item We demonstrate that our multi-agent framework effectively addresses the limitations of existing threshold-dependent methods, providing robust, interpretable, and practically applicable threshold-free classification in diverse real-world detection scenarios.
\end{itemize}

\section{Stylistics and Authorship Attribution}

Stylistics, tracing its intellectual heritage back to classical rhetoric, literary criticism, and early linguistic theory~\cite{aristotle1954rhetoric, jakobson1960clsoing, halliday1978language}, has evolved into an empirically grounded linguistic discipline dedicated to the systematic investigation of textual style. The core objective of stylistics is to identify and analyze the distinctive linguistic patterns—such as lexical selections, syntactic structures, semantic preferences, and rhetorical devices—that constitute a text's unique stylistic fingerprint and reveal insights about its author’s identity, intentions, and cognitive processes~\cite{simpson2004stylistics, short2015style}. These stylistic patterns have been consistently demonstrated to be stable across texts by the same author, making stylistics particularly powerful in authorship attribution tasks~\cite{holmes1994authorship, juola2008authorship}.

\paragraph{Authorship Attribution as a Stylistic Task.}
Authorship attribution, historically rooted in literary scholarship~\cite{mosteller1964inference, burrows2002delta, holmes1994authorship}, employs rigorous stylistic analysis to identify authors by systematically quantifying and comparing linguistic features across texts. Traditionally, this involves detecting cognitively distinctive and stylistically consistent markers unique to individual authors, ranging from lexical and syntactic choices to subtle rhetorical and semantic nuances~\cite{burrows2002delta, stamatatos2009survey}. The emergence of advanced generative AI models, such as GPT-3 and GPT-4~\cite{brown2020language, openai2024gpt4technicalreport}, has expanded the scope of authorship attribution research, necessitating robust stylistic methodologies capable of differentiating between human-authored and AI-generated texts~\cite{ippolito2020automatic, mitchell2023detectgpt}. From this perspective, AI-generated text detection naturally maps onto the theoretical and methodological framework of authorship attribution, becoming a binary classification task distinguishing between \textit{human} and \textit{AI} authorship based on intrinsic stylistic indicators.

\paragraph{Theoretical Foundations: Three Dimensions of Stylistics.}

A substantial body of classical and contemporary stylistics research identifies three foundational linguistic dimensions that collectively constitute textual style~\cite{halliday1978language, simpson2004stylistics, short2015style}. These dimensions provide a comprehensive theoretical framework for analyzing authorship and differentiating between human-written and AI-generated texts. Below, we provide illustrative examples from our sampled stylistic detection guidelines, with the complete set detailed comprehensively in Appendix~\ref{appendix:full_guidelines}:

\begin{showcase}[title=Illustrative Examples of Guidelines]


|\classbg{Semantic Dimension}|
Directness of Claims and Nuanced Expertise  
Human-written: Makes direct, succinct claims without extensive hedging.  
AI-generated: Generally cautious, balanced, or neutral in making claims.
...

|\classbg{Stylistic Dimension}|
Precision and Conciseness  
Human-written: Concise, direct introduction of concepts.  
AI-generated: Balanced, explanatory, and carefully phrased.
...

|\classbg{Structural Dimension}|
Structural Uniformity and Rhythmic Predictability
Human-written: Variable sentence structure and rhythm.  
AI-generated: Uniform, predictable, consistently structured.
...

\end{showcase}

The \textit{semantic dimension}, grounded in Halliday’s systemic functional linguistics~\cite{halliday1978language, halliday2013halliday}, encompasses textual features related to meaning-making, conceptual complexity, thematic originality, and informational specificity. Such semantic traits reflect intricate cognitive processes and intellectual contributions that characterize human writing and differentiate it from mechanically generated text~\cite{halliday1978language, eggins2004introduction}.

The \textit{stylistic dimension}, extensively explored by seminal stylisticians such as Leech, Short, and Simpson~\cite{short2015style, simpson2004stylistics}, addresses expressive nuances, lexical idiosyncrasies, cognitive friction, and subtle linguistic irregularities arising naturally from human cognitive limitations and creativity. These stylistic irregularities have been shown empirically to resist consistent replication by algorithmic processes, thereby serving as crucial discriminative markers in distinguishing between human and AI authorship~\cite{burrows2002delta, stamatatos2009survey, ippolito2020automatic}.

The \textit{structural dimension}, informed by rhetorical and textual coherence theories~\cite{mann1988rhetorical, van1983strategies}, analyzes organizational coherence, discourse arrangement, argumentative structures, and patterns of logical progression. Human authors typically display sophisticated and contextually adaptive coherence strategies and nuanced argumentation structures. Conversely, AI-generated texts frequently exhibit repetitive organizational patterns, predictable coherence mechanisms, and shallow argumentative structures, revealing their algorithmic origins~\cite{gehrmann2019gltr, mitchell2023detectgpt}.

\section{Discussion on Evaluation Metrics}
\label{sec:evaluation_metrics}

In this study, we primarily employ \textbf{Accuracy} as our central evaluation metric, supplemented by \textbf{Area Under the Receiver Operating Characteristic (AUROC)} as a secondary reference. This decision is motivated by practical considerations and intrinsic methodological constraints inherent to prompt-based Large Language Model (LLM) approaches for AI-generated text detection.

\paragraph{Accuracy} measures the proportion of correctly classified samples to the total number of samples, naturally aligning with real-world requirements for clear, actionable binary decisions. Existing detection methods~\citep{mitchell2023detectgpt, bao2024fastdetectgptefficientzeroshotdetection} frequently rely on metrics like AUROC, which, while theoretically informative, require selecting optimal thresholds based on external, domain-specific calibration datasets. In practical scenarios, such datasets are often unavailable, significantly limiting these methods’ applicability and scalability. In contrast, our proposed method uniquely employs a threshold-free aggregation strategy, eliminating dependency on external calibration sets, thereby making Accuracy an inherently more practical and relevant metric for evaluation.

\paragraph{AUROC} provides a comprehensive measure of detection performance across all possible thresholds, quantifying the probability that a randomly selected AI-generated text receives a higher predicted probability of being machine-generated compared to a randomly chosen human-written text. AUROC scores range between 0.0 (completely incorrect classification) and 1.0 (perfect classification), with 0.5 indicating random guessing. However, recent studies have emphasized the limitations of AUROC for evaluating prompt-based LLM classifiers, highlighting the inherently skewed and poorly calibrated probability distributions produced by generative LLMs~\citep{giovannotti2024calibratedlargelanguagemodels, jiang2023calibrating, yoon2025usingtournamentscalculateauroc}. Prompt-based methods typically yield discrete predictions and uncalibrated confidence scores, rendering direct AUROC comparisons with traditional discriminative classifiers intrinsically disadvantageous and potentially misleading~\citep{zhang2024don}.

Consequently, we emphasize Accuracy as our primary evaluation metric due to its direct applicability and practical advantages, while reporting AUROC as an auxiliary measure to demonstrate that our approach remains competitively robust even under evaluation criteria less favorable to threshold-free methods.

\section{Methodology}

Our proposed multi-agent framework operationalizes classical stylistic theory to achieve robust zero-shot detection of AI-generated texts without relying on labeled datasets or external threshold tuning. The framework consists of three main components: a \textit{router agent} for adaptive activation of guideline-specific \textit{base agents}, guideline-specific base agents organized into three stylistic dimensions (\textit{semantic}, \textit{stylistic}, and \textit{structural}), and a \textit{meta agent} synthesizing these analyses into a final interpretable classification. All agents are instantiated via Large Language Models (LLMs) and utilize the advanced semantic-steering confidence calibration method (\textit{SteeringConf})~\cite{zhou2025calibratingllmconfidencesemantic}.

\subsection{Router Agent}

The router agent determines which specific stylistic guidelines—and consequently, which corresponding base agents—are activated for evaluating the input text. It achieves this by first analyzing each input text using an LLM to jointly infer its topical domain (e.g., medical, legal, literary) and stylistic attributes (e.g., formal register, argumentative coherence). Based on this inferred textual profile, the router agent selectively activates a subset of guideline-specific base agents within each linguistic dimension. For instance, when analyzing a medical abstract characterized by formal language, guidelines emphasizing semantic clarity and structural precision are activated, thus activating their corresponding base agents, while guidelines related to emotional expressiveness and their corresponding agents remain inactive.

\subsection{Base Agents}

Each activated guideline-specific base agent independently evaluates the input text according to its assigned stylistic guideline using a Large Language Model (LLM). For each input text, the base agent initially produces three foundational outputs:

\begin{itemize}[leftmargin=*, itemsep=2pt]
    \item A binary classification decision (\textit{AI-generated} or \textit{human-written}) based on the given guideline.
    \item An explicit textual rationale highlighting linguistic evidence supporting this decision, aligned with dimension-specific stylistic theory.
    \item A verbalized confidence score indicating the LLM’s certainty in the decision.
\end{itemize}

To robustly calibrate these confidence scores, we further employ the advanced semantic-steering confidence calibration method (\textit{SteeringConf}). Specifically, the base agent independently evaluates the same input text using five symmetric steering prompts—\textit{very cautious}, \textit{cautious}, \textit{vanilla}, \textit{confident}, and \textit{very confident}. Under each steering prompt \(P_k \in \mathcal{P}\), the agent produces a binary classification prediction \(f_k(x)\), an explicit textual rationale justifying this prediction, and an associated verbalized confidence score \(c_k(x)\).

These multiple prompt-based outputs are then aggregated into a single robust calibrated confidence \( c_\text{cal}(x) \) by integrating two key consistency metrics:

\textit{Answer Consistency} (\(\kappa_{\mathrm{ans}}\)): measures prediction agreement across prompts:
\[
\kappa_{\mathrm{ans}} = \frac{1}{|\mathcal P|}\,\max_{y\in\{\text{AI},\text{Human}\}}\sum_k \mathbb{I}[f_k(x)=y]
\]

\textit{Confidence Consistency} (\(\kappa_\text{conf}\)): evaluates confidence stability across prompts:
\[
\resizebox{\columnwidth}{!}{$
  \mu_c = \frac{1}{|\mathcal P|}\sum_{k} c_k(x),\quad
  \sigma_c = \sqrt{\frac{1}{|\mathcal P|}\sum_{k}(c_k(x)-\mu_c)^2}
$}
\]
\[
\kappa_\text{conf} = \frac{1}{1+\sigma_c/\mu_c}
\]

These metrics are combined multiplicatively with the average confidence \(\mu_c\) to yield the final calibrated confidence score:
\[
c_\text{cal}(x) = \mu_c \cdot \kappa_\text{ans} \cdot \kappa_\text{conf}
\]

Finally, the guideline-specific base agent’s ultimate classification decision \(f_{\mathrm{final}}(x)\) is selected from the steering prompt whose original confidence \(c_k(x)\) most closely matches this calibrated confidence \( c_\text{cal}(x) \):
\[
\resizebox{\columnwidth}{!}{$
  k^* = \operatorname*{arg\,min}_{k}\bigl|c_k(x)-c_{\mathrm{cal}}(x)\bigr|,\quad
  f_{\mathrm{final}}(x)=f_{k^*}(x)
$}
\]
The calibrated classification decision, rationale, and confidence constitute the final structured \textit{guideline-specific report}.

\subsection{Meta Agent}
The meta agent synthesizes the guideline-specific reports from all activated base agents through a confidence-weighted aggregation process. Specifically, it integrates the classification decisions, robustly calibrated confidence scores, and explicit textual rationales provided by each guideline-specific agent. The meta agent dynamically evaluates and prioritizes agent inputs according to their confidence scores and the quality of linguistic reasoning presented. When agents disagree, it contextually assesses and reconciles differing rationales, forming a nuanced, consensus-based final classification. Subsequently, it generates a consolidated textual rationale summarizing the critical linguistic evidence underlying the final aggregated decision, alongside an aggregated confidence score computed using the same semantic-steering calibration mechanism employed by the base agents.

\section{Experimental Setup}

\subsection{Datasets}

We adopt publicly available datasets from \citet{bao2024fastdetectgptefficientzeroshotdetection}, which closely adhere to the experimental settings established by \citet{mitchell2023detectgpt}. Specifically, these datasets contain texts generated by three widely-used commercial closed-source language models: ChatGPT (gpt-3.5-turbo)\footnote{\url{https://platform.openai.com/docs/models/gpt-3-5-turbo}}, GPT-4 (gpt-4-0613)\footnote{\url{https://platform.openai.com/docs/models/gpt-4-turbo-and-gpt-4}}, and Claude-3 Sonnet (claude-3-sonnet-20240229) and Opus (claude-3-opus-20240229)\footnote{\url{https://docs.anthropic.com/en/docs/models-overview}}. We refer to these models collectively as the \emph{source models} for AI-generated texts.

Consistent with prior work~\cite{bao2024fastdetectgptefficientzeroshotdetection}, our evaluation focuses on three distinct datasets: \textit{XSum}\cite{narayan2018dontdetailsjustsummary} for news article summarization, \textit{WritingPrompts}\cite{fan2018hierarchical} for creative story generation, and \textit{PubMedQA}~\cite{jin2019pubmedqa} for biomedical research question answering. These datasets provide diverse textual contexts to comprehensively assess the generalizability of detection methods. Each dataset comprises original human-written texts and corresponding texts generated by the aforementioned source models, collectively forming the benchmark dataset used in our evaluation.

\subsection{Implementation Details}
For all agents—including the router agent, base agents, and meta agent—we employed \texttt{deepseek-chat-v3-0324}~\cite{deepseekai2025deepseekv3technicalreport} as the base Large Language Model, due to its demonstrated robustness and reliability in interpretation and reasoning tasks. The complete set of prompts utilized in our experiments is provided in Appendix\ref{appendix:prompts}.

\subsection{Baselines}
For supervised methods, we benchmark against GPT-2 detectors based on RoBERTa-base and RoBERTa-large models~\cite{liu2019roberta}, as well as GPTZero~\cite{tian2023gptzero}. For zero-shot detectors, our comparisons include recent advanced approaches such as Fast-DetectGPT~\cite{bao2024fastdetectgptefficientzeroshotdetection}, DetectGPT~\cite{mitchell2023detectgpt}, NPR~\cite{su2023detectllm}, and DNA-GPT~\cite{yang2023dna}. Additionally, we include classic zero-shot techniques frequently utilized in previous literature, specifically Likelihood (mean log probabilities)\cite{gehrmann2019gltr}, LogRank (average log-ranks of token probabilities)\cite{solaiman2019release}, Entropy (mean token entropy of predictive distributions)\cite{ippolito2020automatic}, and LRR (a combined metric of log probability and log-rank)\cite{su2023detectllm}. Detailed experimental settings for these baseline methods are provided in Appendix~\ref{sec:baseline_details}.

\section{Results}

\begin{table*}[h]
\centering
\small
\setlength{\tabcolsep}{7pt}
\renewcommand{\arraystretch}{1.035}
\begin{tabular}{lcccccccc}
\toprule
\multirow{2}{*}{\textbf{Method}} 
  & \multicolumn{4}{c}{\textbf{ChatGPT (gpt-3.5-turbo)}} 
  & \multicolumn{4}{c}{\textbf{GPT-4}} \\
\cmidrule(lr){2-5} \cmidrule(lr){6-9}
 & XSum & Writing & PubMed & Avg. & XSum & Writing & PubMed & Avg. \\
\midrule
RoBERTa-base     & 0.8367 & 0.6000 & 0.5267 & 0.6545 & 0.6242 & 0.4967 & 0.4396 & 0.5202 \\
RoBERTa-large    & 0.6567 & 0.5033 & 0.4800 & 0.5467 & 0.5336 & 0.4933 & 0.4732 & 0.5000 \\
\midrule
Likelihood (Neo-2.7)      & 0.8000 & 0.9133 & \textbf{0.7633} & \underline{0.8256} 
                 & 0.7100 & 0.6067 & \underline{0.7233} & \underline{0.6800} \\
Entropy (Neo-2.7)         & 0.3800 & 0.3767 & 0.3533 & 0.3700 
                 & 0.5000 & 0.4767 & 0.3967 & 0.4578 \\
Rank (Neo-2.7)            & 0.6767 & 0.6067 & 0.5800 & 0.6211 
                 & 0.6000 & 0.5100 & 0.5767 & 0.5622 \\
LogRank (Neo-2.7)         & 0.8300 & 0.8900 & 0.7533 & 0.8244 
                 & \underline{0.7267} & 0.5900 & 0.7200 & 0.6789 \\
LRR (Neo-2.7)             & 0.8433 & 0.7300 & 0.5767 & 0.7167 
                 & 0.6846 & 0.5467 & 0.5570 & 0.5961 \\
DNA-GPT (Neo-2.7)         & 0.7100 & 0.8467 & 0.5000 & 0.6856 
                 & 0.6600 & 0.7100 & 0.5000 & 0.6233 \\
NPR (T5-11B/Neo-2.7)             & 0.7033 & 0.8867 & 0.6000 & 0.7300 
                 & 0.4966 & 0.6100 & 0.5537 & 0.5534 \\
DetectGPT (T5-11B/Neo-2.7)       & 0.6467 & 0.7667 & 0.4967 & 0.6367 
                 & 0.5067 & 0.5033 & 0.5000 & 0.5033 \\
Fast-Detect (GPT-J/Neo-2.7)  & \textbf{0.9467} & \textbf{0.9333} & 0.5433 & 0.8078 
                 & 0.7100 & \underline{0.7600} & 0.5267 & 0.6656 \\
\midrule
AGENT-X (Ours)   & \underline{0.8967} & \underline{0.9233} & \underline{0.7604} & \textbf{0.8601} 
                 & \textbf{0.8624} & \textbf{0.8705} & \textbf{0.8446} & \textbf{0.8592} \\
\bottomrule
\end{tabular}
\caption{Detection of \emph{ChatGPT (gpt-3.5-turbo)} and \emph{GPT-4} generations, evaluated in accuracy. Thresholds for baseline methods are optimized on a separate validation set (SQuAD-gpt-neo-2.7B dataset). Highest in each column is \textbf{bold}, second-highest is \underline{underlined}.}
\label{tab:chatgpt_gpt4_results_acc}
\end{table*}

\begin{table*}[h]
\centering
\small
\setlength{\tabcolsep}{7pt}
\renewcommand{\arraystretch}{1.035}
\begin{tabular}{lcccccccc}
\toprule
\multirow{2}{*}{\textbf{Method}} 
  & \multicolumn{4}{c}{\textbf{Claude-3-Opus}} 
  & \multicolumn{4}{c}{\textbf{Claude-3-Sonnet}} \\
\cmidrule(lr){2-5} \cmidrule(lr){6-9}
 & XSum & Writing & PubMed & Avg. & XSum & Writing & PubMed & Avg. \\
\midrule
RoBERTa-base   & 0.7967 & 0.6300 & 0.4333 & 0.6200 
               & 0.7000 & 0.5700 & 0.4433 & 0.5711 \\
RoBERTa-large  & 0.6600 & 0.5133 & 0.4367 & 0.5367 
               & 0.5867 & 0.4900 & 0.4333 & 0.5033 \\
\midrule
Likelihood (Neo-2.7)    & 0.8033 & 0.9000 & 0.7500 & 0.8178 
               & 0.7833 & 0.8367 & \underline{0.7367} & \underline{0.7856} \\
Entropy (Neo-2.7)       & 0.4233 & 0.3367 & 0.3800 & 0.3800 
               & 0.4667 & 0.3900 & 0.3733 & 0.4100 \\
Rank (Neo-2.7)          & 0.6600 & 0.6467 & 0.5733 & 0.6267 
               & 0.6200 & 0.6133 & 0.5833 & 0.6056 \\
LogRank (Neo-2.7)       & 0.8167 & 0.8867 & \underline{0.7567} & \underline{0.8200} 
               & \underline{0.7933} & 0.8200 & 0.7233 & 0.7789 \\
LRR (Neo-2.7)           & 0.8167 & 0.8100 & 0.6333 & 0.7533 
               & 0.7367 & 0.7333 & 0.6567 & 0.7089 \\
DNA-GPT (Neo-2.7)       & \underline{0.8200} & 0.8667 & 0.5000 & 0.7289 
               & 0.7733 & 0.8467 & 0.5000 & 0.7067 \\    
NPR (T5-11B/Neo-2.7)    & 0.7400 & \underline{0.9233} & 0.6900 & 0.7844 
               & 0.7333 & \textbf{0.9067} & 0.6800 & 0.7733 \\
DetectGPT (T5-11B/Neo-2.7) & 0.5000 & 0.5000 & 0.6567 & 0.5522 
               & 0.5000 & 0.5000 & 0.6567 & 0.5522 \\
Fast-Detect (GPT-J/Neo-2.7) & \textbf{0.8767} & \textbf{0.9400} & 0.5700 & 0.7956 
               & \textbf{0.8067} & \underline{0.8700} & 0.5300 & 0.7356 \\
\midrule
AGENT-X (Ours) & 0.8133 & 0.9100 & \textbf{0.7900} & \textbf{0.8378} 
               & 0.7567 & 0.8633 & \textbf{0.7800} & \textbf{0.8000} \\
\bottomrule
\end{tabular}
\caption{Detection of \emph{Claude-3-Opus} and \emph{Claude-3-Sonnet} generations, evaluated in accuracy. Thresholds for baseline methods are optimized on a separate validation set (SQuAD-gpt-neo-2.7B dataset). Highest values in each column are \textbf{bolded}, second-highest are \underline{underlined}.}
\label{tab:claude3opus_sonnet_results_acc}
\end{table*}

Table~\ref{tab:chatgpt_gpt4_results_acc} compares the accuracy (ACC) of our \textit{AGENT-X} framework with supervised and zero-shot baselines for texts generated by ChatGPT and GPT-4 across three datasets (\textit{XSum}, \textit{Writing}, \textit{PubMed}). Overall, \textit{AGENT-X} achieves the highest average accuracy (\textbf{86.0\%} ChatGPT, \textbf{85.9\%} GPT-4), significantly surpassing the best zero-shot baselines (\textit{Likelihood}: 82.6\% ChatGPT, 68.0\% GPT-4; \textit{Fast-DetectGPT}: 80.8\% ChatGPT, 66.6\% GPT-4). While \textit{Fast-DetectGPT} performs strongly in specific scenarios (e.g., \textit{XSum}, \textit{Writing} for ChatGPT), it generalizes poorly to biomedical texts (\textit{PubMed}), highlighting the superior domain adaptability of our threshold-free method.

Table~\ref{tab:claude3opus_sonnet_results_acc} presents accuracy comparisons on \emph{Claude-3} datasets. \textit{AGENT-X} consistently achieves the highest average accuracy (\textbf{83.8\%} for Opus, \textbf{80.0\%} for Sonnet), notably outperforming zero-shot methods including \textit{Likelihood} and \textit{Fast-DetectGPT}. Particularly on \textit{PubMed}, \textit{AGENT-X} delivers exceptional performance (\textbf{79.0\%} Opus, \textbf{78.0\%} Sonnet), underscoring its robustness and practical applicability across diverse real-world contexts.

We additionally evaluated our framework using AUROC metrics, with comprehensive results provided in Appendix~\ref{appendix:additional_results}. Although AUROC metrics inherently favor threshold-dependent methods and thus are less aligned with our threshold-free design, \textit{AGENT-X} consistently exhibits strong and competitive performance across all evaluated datasets and language models. This highlights our framework’s robustness and reliability, even under evaluation settings that do not fully leverage its advantages.

\begin{figure*}[h]
\centering
\includegraphics[width=1\linewidth]{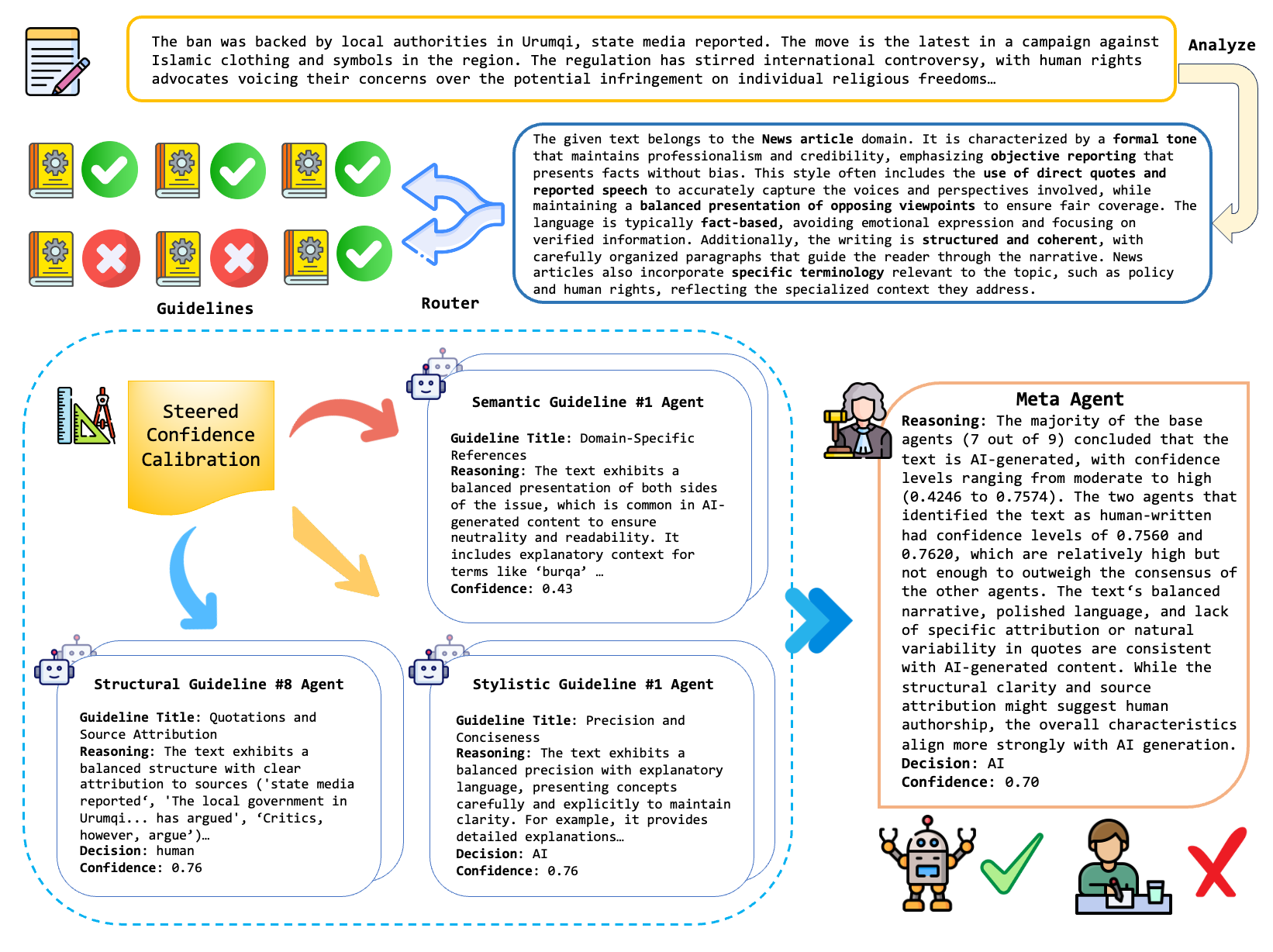}
\caption{Example detection case of \textit{AGENT-X}. The router agent dynamically activates relevant stylistic guidelines based on inferred textual characteristics. Each guideline-specific agent independently assesses the input text, providing calibrated confidence scores and detailed linguistic reasoning. The meta agent aggregates these evaluations via confidence-weighted voting, making the final interpretable classification decision.}
\label{fig:case}
\end{figure*}

We further demonstrate the interpretability and effectiveness of our proposed framework through an illustrative detection example shown in Figure~\ref{fig:case}.

\begin{table*}[h]
\centering
\small
\setlength{\tabcolsep}{8pt}
\renewcommand{\arraystretch}{1.035}
\begin{tabular}{lcccccccc}
\toprule
\multirow{2}{*}{\textbf{Method}} 
  & \multicolumn{4}{c}{\textbf{ACC}} 
  & \multicolumn{4}{c}{\textbf{AUROC}} \\
\cmidrule(lr){2-5} \cmidrule(lr){6-9}
 & XSum & Writing & PubMed & Avg. & XSum & Writing & PubMed & Avg. \\
\midrule
Full \textbf{AGENT-X} 
               & \underline{0.8624} & \textbf{0.8705} & \textbf{0.8446} & \textbf{0.8592} 
               & \textbf{0.9367} & \textbf{0.9206} & \textbf{0.8447} & \textbf{0.9007} \\
w/o Multi-Agent 
               & \textbf{0.8691} & 0.7967 & 0.6946 & 0.7868 
               & \underline{0.9357} & 0.8674 & \underline{0.8323} & \underline{0.8785} \\
w/o Guidelines 
               & 0.7785 & 0.5633 & 0.5000 & 0.6139 
               & 0.9319 & 0.7078 & 0.7078 & 0.7825 \\
w/o LLM Router 
               & 0.8054 & 0.8067 & 0.5470 & 0.7197 
               & 0.8652 & 0.8341 & 0.6221 & 0.7738 \\
w/o Adaptive Routing 
               & \textbf{0.8691} & \underline{0.8267} & 0.5805 & 0.7588 
               & 0.9236 & 0.8854 & 0.7906 & 0.8665 \\
w/o Steer Calibration 
               & 0.8590 & 0.8100 & \underline{0.7282} & \underline{0.7991} 
               & 0.9144 & \underline{0.8914} & 0.7383 & 0.8480 \\
\bottomrule
\end{tabular}
\caption{Ablation study results demonstrating the performance impact of removing or modifying key components of AGENT-X on GPT-4–generated datasets. Highest values in each column are \textbf{bolded}, second-highest are \underline{underlined}.}
\label{tab:ablation_study}
\end{table*}

\section{Ablation Study}
We conducted an ablation study to rigorously validate the contribution of each critical component in the \textit{AGENT-X} framework. Specifically, we systematically removed or altered the following components: (1)\textit{Multi-Agent}, (2)\textit{Guidelines}, (3)\textit{LLM Router}, (4)\textit{Adaptive Routing Strategy}, and (5)\textit{Steer-based Confidence Calibration}. Performance was evaluated primarily using Accuracy (\textbf{ACC}), supplemented by AUROC for additional context (Table\ref{tab:ablation_study}).

\paragraph{w/o Multi-Agent.}
Replacing the specialized dimension-specific linguistic agents with a single comprehensive agent led to a notable decline in average accuracy (from 85.9\% to 78.7\%). This emphasizes the critical advantage provided by having dedicated agents, each conducting focused analyses aligned with distinct linguistic dimensions.

\paragraph{w/o Guidelines.}
Eliminating structured detection guidelines and replacing them with randomly selected prompts resulted in the most significant degradation in average accuracy (down to 61.4\%), underscoring the essential role that theoretically-grounded detection guidelines play in enabling interpretable and effective detection.

\paragraph{w/o LLM Router.}
Replacing the intelligent guideline-selection mechanism of the router agent with random guideline activation significantly reduced accuracy (from 85.9\% to 72.0\%). This highlights the crucial adaptive capability provided by intelligently selecting guidelines based on inferred textual characteristics.

\paragraph{w/o Adaptive Routing (Full Guidelines).}
Activating all available guidelines simultaneously, rather than selectively, resulted in moderate performance degradation (average accuracy dropped to 75.9\%). This indicates that excessive guideline activation introduces unnecessary noise and complexity, reinforcing the value of an adaptive, context-sensitive routing strategy.

\paragraph{w/o Steer-based Calibration.}
Using uncalibrated confidence scores directly from LLMs, rather than applying our semantic-steering calibration method, decreased average accuracy from 85.9\% to 79.9\%. This clearly demonstrates the effectiveness of semantic steering calibration in enhancing confidence reliability and overall detection accuracy.

\section{Related Work}

\paragraph{Confidence Estimation} 
Reliable confidence estimation remains critical yet challenging for the practical deployment of Large Language Models (LLMs). Existing methods primarily leverage external knowledge~\citep{tsai2024efficient,mielke2022reducing}, internal logits~\citep{jiang2021how,tian2023just}, or direct verbalized assessments~\citep{lin2022teaching,kadavath2022language}. Recent integrated frameworks, notably multi-prompt semantic steering (\textit{SteeringConf})~\citep{zhou2025calibratingllmconfidencesemantic}, systematically manipulate prompts to enhance calibration robustness. Building upon this line of research, our approach employs semantic steering-based confidence estimation, improving interpretability and reliability in zero-shot detection of AI-generated texts.

\paragraph{AI-generated Text Detection}
Detecting AI-generated texts, especially in zero-shot settings, is increasingly challenging. Black-box approaches like DetectGPT~\citep{mitchell2023detectgpt}, NPR~\citep{su2023detectllm}, and DNA-GPT~\citep{yang2023dna} typically involve multiple evaluations or generations, incurring high computational overhead. White-box methods utilize model-internal metrics such as entropy, likelihood, and log-rank~\citep{gehrmann2019gltr,bao2024fastdetectgptefficientzeroshotdetection,tulchinskii2024intrinsic}. Meanwhile, supervised classifiers~\citep{bakhtin2019real,yan2023detection,verma2024ghostbuster} and watermarking approaches~\citep{kirchenbauer2023watermark,zhao2023provable} rely either on extensive training data or proactive intervention during text generation, limiting their applicability with proprietary models. Unlike these methods, our proposed framework introduces a robust, interpretable, and threshold-free zero-shot detection mechanism, eliminating reliance on labeled datasets and external calibration.

\paragraph{LLM Multi-Agent Systems}
Large Language Model Multi-Agent Systems (LLM-MAS) exploit collective intelligence through collaboration among multiple specialized agents~\citep{guo2024large,han2024llm}. These systems commonly utilize structured communication protocols such as debates, voting, and task-specific dialogues~\citep{du2023improving,zhao2024electoral,hong2023metagpt}, demonstrating effectiveness across various domains including software development~\citep{qian2024chatdev}, embodied agents~\citep{song2023llm}, and simulations~\citep{park2023generative,wang2023avalon}. Inspired by this paradigm, we propose a novel LLM-based multi-agent framework that dynamically integrates specialized linguistic analyses, significantly enhancing adaptability and generalization for zero-shot AI-generated text detection.

\section{Conclusion}
In this work, we introduced a novel zero-shot multi-agent collaboration framework for robust detection of AI-generated texts, grounded in systematically structured stylistic theory. Our method leverages specialized linguistic agents, advanced semantic steering-based confidence calibration, and a dynamic Mixture-of-Agent expert routing mechanism to deliver state-of-the-art detection performance. Experimental results demonstrated substantial improvements in interpretability, robustness, and generalizability over existing supervised and zero-shot detection methods. Our approach provides a practical, threshold-free solution, significantly advancing AI-generated text detection in real-world zero-shot scenarios.

\section*{Limitations}
Although our framework demonstrates strong performance and interpretability across diverse contexts, it requires relatively capable base language models for effective agent instantiation. Employing less powerful or smaller-scale LLMs could pose challenges in comprehensively understanding the detection guidelines or accurately performing agent-specific tasks.

\section*{Ethical Considerations}
Our proposed framework contributes positively by facilitating transparent and interpretable detection of AI-generated texts, addressing ethical concerns around misinformation, authenticity, and accountability. However, careful consideration should be given to potential biases inherent in the underlying base language models, as these biases could inadvertently influence agent decisions. We encourage users and researchers to employ our framework responsibly, being mindful of potential impacts on fairness, privacy, and transparency in various application contexts.

\bibliography{custom}

\newpage
\appendix

\section{Detection Guidelines}
\label{appendix:full_guidelines}
\begin{showcase}[title=Detection Guidelines]


|\classbg{Semantic Dimension}|

|\classbg{Guideline 1}|: Domain-Specific References  
Human-written: Briefly mentions specialized terms without explanation.  
AI-generated: Often explicitly clarifies or cautiously explains specialized terms.

|\classbg{Guideline 2}|: Directness of Claims and Nuanced Expertise  
Human-written: Makes direct, succinct claims without extensive hedging.  
AI-generated: Generally cautious, balanced, or neutral in making claims.

|\classbg{Guideline 3}|: Explanatory or Clarifying Intent  
Human-written: Reports specific findings or outcomes without broad explanations.  
AI-generated: Frequently clarifies or explains general concepts or terminologies.

|\classbg{Guideline 4}|: Specific Names, Places, and Contextual Details  
Human-written: Frequently contains quirky, casual, or humorously exaggerated specifics.  
AI-generated: Usually general or predictable specifics, lacking genuine spontaneity.

|\classbg{Guideline 5}|: Realistic Variability in Detail and Attribution  
Human-written: Contextually appropriate and varied attribution of statements.  
AI-generated: Uniform, repetitive, or artificially formalized attribution.

|\classbg{Guideline 6}|: Authenticity and Originality of Commentary  
Human-written: Includes authentic, clearly attributed commentary.  
AI-generated: Generic, balanced commentaries lacking authentic sourcing.

|\classbg{Stylistic Dimension}|

|\classbg{Guideline 1}|: Precision and Conciseness  
Human-written: Concise, direct introduction of concepts.  
AI-generated: Balanced, explanatory, and carefully phrased.

|\classbg{Guideline 2}|: Immediate Use of Abbreviations after Specialized Terms  
Human-written: Usually does not immediately pair terms with abbreviations.  
AI-generated: Occasionally explicitly pairs specialized terms with abbreviations (e.g., 'Term (ABBR)').

|\classbg{Guideline 3}|: Conversational or QA-Like Openings  
Human-written: Rarely uses conversational openings ("Yes," "No,").  
AI-generated: Occasionally uses explicit conversational openings, strongly signaling AI authorship.

|\classbg{Guideline 4}|: Metaphors and Rhetorical Expression  
Human-written: Subtle humor, ironic undertones, emotionally nuanced metaphors.  
AI-generated: Clear, systematically integrated metaphors lacking spontaneity.

|\classbg{Guideline 5}|: Unusual Vocabulary, Punctuation, and Expressive Conventions  
Human-written: Often contains informal, exaggerated punctuation or stylistic quirks.  
AI-generated: Uses conventional vocabulary and formal, standardized punctuation.

|\classbg{Guideline 6}|: Clarity of Exposition and Intent  
Human-written: Can be ambiguous or nuanced in intention or meaning.  
AI-generated: Maintains explicit clarity, coherence, and straightforward intent.

|\classbg{Guideline 7}|: Stylistic Naturalness vs. Formality  
Human-written: Spontaneously fluctuates between formal and informal styles.  
AI-generated: Consistently formal with minimal stylistic fluctuation.

|\classbg{Guideline 8}|: Balanced and Contextually Embedded Narratives  
Human-written: Naturally embeds quotations and realistic perspectives.  
AI-generated: Artificially balanced narratives with formulaic phrasing.

|\classbg{Guideline 9}|: Realistic Variability in Quotation  
Human-written: Quotes vary realistically in length, specificity, and language.  
AI-generated: Uniformly polished, generic, or repetitive quotations.

|\classbg{Guideline 10}|: Repetitive, Generic, or Overly Balanced Wording  
Human-written: Uses varied and contextually nuanced phrasing.  
AI-generated: Frequently employs generic or formulaic rhetorical phrases.

|\classbg{Structural Dimension}|

|\classbg{Guideline 1}|: Use of Explicit Logical Connectors  
Human-written: Rarely uses explicit logical connectors; context-dependent.  
AI-generated: Slightly more frequent explicit logical connectors ("Therefore," "However,").

|\classbg{Guideline 2}|: Structural Uniformity and Rhythmic Predictability
Human-written: Variable sentence structure and rhythm.  
AI-generated: Uniform, predictable, consistently structured.

|\classbg{Guideline 3}|: Emotional Nuance and Narrative Flow  
Human-written: Subtle emotional nuances, spontaneous and uneven flow.  
AI-generated: Consistent emotional portrayal, purpose-driven and evenly structured narrative.

|\classbg{Guideline 4}|: Plot and Detail Distribution  
Human-written: Includes tangential, abrupt, or uneven narrative details.  
AI-generated: Evenly distributed, explicitly connected plot details.

|\classbg{Guideline 5}|: Narrative Structure Consistency  
Human-written: Uneven structure, sudden shifts, unresolved narrative threads.  
AI-generated: Methodical, evenly balanced narrative structure.

|\classbg{Guideline 6}|: Narrative Detail and Exposition  
Human-written: Incidental, loosely connected details, spontaneous digressions.  
AI-generated: Strictly relevant, methodically structured narrative details.

|\classbg{Guideline 7}|: Use of Dialogue  
Human-written: Dialogues include natural irregularities or emotional nuances.  
AI-generated: Dialogues explicitly reinforce plot, appear structured and formal.

|\classbg{Guideline 8}|: Quotations and Source Attribution  
Human-written: Clearly attributed authentic quotations from named sources.  
AI-generated: Generic, polished quotations without realistic nuance.

|\classbg{Guideline 9}|: Structural Clarity and Factual Conciseness  
Human-written: Concise factual delivery with succinct background.  
AI-generated: Excessive explanatory detail or overly comprehensive background.


\end{showcase}

\section{Detection Case}
\begin{showcase}[title=Detection Case]

|\classbg{Input Text}|

The ban was backed by local authorities in Urumqi, state media reported. The move is the latest in a campaign against Islamic clothing and symbols in the region. The regulation has stirred international controversy, with human rights advocates voicing their concerns over the potential infringement on individual religious freedoms. \n\n The local government in Urumqi, the capital city of Xinjiang Uyghur Autonomous Region, has argued that the ban seeks to counteract extremism and maintain social harmony. They insist it is not an attempt to suppress the Islamic faith, but rather a measure to prevent the spread of radical ideologies. The ban includes items such as the burqa, a full-body veil worn by some Muslim women, and \"jilbabs,\" outer garments covering the body.\n\n Critics, however, argue that the ban is a symptom of the broader issue of religious discrimination, particularly against the Uyghur Muslim minority. They warn against the potential for increased social alienation and unrest as a result of such measures

|\classbg{Zero-Shot Domain and Stylistic Feature Inference}|

{
  "Domain": "News article",
  "Stylistic Features": [
    "Formal tone",
    "Objective reporting",
    "Use of direct quotes and reported speech",
    "Balanced presentation of opposing viewpoints",
    "Fact-based language with minimal emotional expression",
    "Structured and coherent paragraphing",
    "Use of specific terminology related to policy and human rights"
  ]
}


|\classbg{Adaptive Guideline Selection}|
Activated Guidelines:
Semantic Guideline 1,2,6
Stylistic Guideline 1,8,9,10
Structural Guideline 8,9
    



|\classbg{Base Agent}|

|\classbg{Semantic Guideline 1 Agent}|
Guideline Title: Domain-Specific References
Reasoning: The text exhibits a balanced presentation of both sides of the issue, which is common in AI-generated content to ensure neutrality and readability. It includes explanatory context for terms like 'burqa' and 'jilbabs,' which AI tends to do to ensure clarity. The structure is coherent and well-organized, typical of AI-generated texts that aim for comprehensive coverage without assuming prior knowledge. The mention of 'Xinjiang Uyghur Autonomous Region' and the detailed explanation of the local government's stance also suggest an AI's tendency to provide background information.
Decision: AI  
Confidence: 0.43


|\classbg{Semantic Guideline 2 Agent}|
Guideline Title: Directness of Claims and Nuanced Expertise
Reasoning: The text exhibits a balanced and explanatory tone, which is characteristic of AI-generated content. It provides a detailed overview of the situation, including perspectives from both the local government and critics, without making abrupt or bold statements. The language is neutral and avoids direct, confident assertions, which aligns with the tendency of AI to hedge and offer balanced claims. Specific examples include the detailed explanation of the ban's purpose and the inclusion of various viewpoints without leaning heavily towards one side.
Decision: AI  
Confidence: 0.69  


|\classbg{Semantic Guideline 6 Agent}|
Guideline Title: Authenticity and Originality of Commentary
Reasoning: The text exhibits a balanced presentation of views, with statements like 'The local government in Urumqi... has argued that the ban seeks to counteract extremism and maintain social harmony' and 'Critics, however, argue that the ban is a symptom of the broader issue of religious discrimination.' These statements are generalized and lack specific attribution to credible sources, which is characteristic of AI-generated content. Additionally, the commentary is neutral and attempts to present both sides without clear, authentic sourcing, further suggesting AI generation.
Decision: AI  
Confidence: 0.72


|\classbg{Stylistic Guideline 1 Agent}|
Guideline Title: Precision and Conciseness
Reasoning: The text exhibits a balanced precision with explanatory language, presenting concepts carefully and explicitly to maintain clarity. For example, it provides detailed explanations about the ban's purpose and the items it includes, such as the burqa and 'jilbabs.' It also includes perspectives from both the local government and critics, which is typical of AI-generated texts aiming for a comprehensive overview. The language is formal and avoids personal anecdotes or emotional expressions, further suggesting an AI origin.
Decision: AI  
Confidence: 0.76


|\classbg{Stylistic Guideline 8 Agent}|
Guideline Title: Balanced and Contextually Embedded Narratives
Reasoning: The text exhibits a balanced narrative, presenting both the local government's perspective and the critics' viewpoint. However, the balance feels somewhat formulaic and predictable, with phrases like 'stirred international controversy' and 'argued that the ban seeks to counteract extremism and maintain social harmony' appearing overly polished and structured. The acknowledgment of multiple perspectives is done in a manner that feels more like a template than a spontaneous, nuanced discussion. This suggests an AI-generated text rather than a human-written one.
Decision: AI  
Confidence: 0.71

|\classbg{Stylistic Guideline 9 Agent}|
Guideline Title: Realistic Variability in Quotation
Reasoning: The text exhibits a high level of polish and formality, with a consistent tone throughout. The quotes and paraphrased statements are uniformly detailed and lack the casual variability typical of human writing. For example, the phrases 'The move is the latest in a campaign against Islamic clothing and symbols in the region' and 'They insist it is not an attempt to suppress the Islamic faith, but rather a measure to prevent the spread of radical ideologies' are overly formal and lack the natural variability in length and specificity seen in human speech. Additionally, the text lacks the occasional incomplete or casually phrased sentences that would indicate human authorship.
Decision: AI  
Confidence: 0.58

|\classbg{Stylistic Guideline 10 Agent}|
Guideline Title: Repetitive, Generic, or Overly Balanced Wording
Reasoning: The text exhibits a balanced and structured narrative, with phrases like 'stirred international controversy' and 'maintain social harmony,' which are characteristic of AI-generated content. The language is polished and lacks the natural imperfections or unique expressions typically found in human writing. Additionally, the systematic presentation of arguments for and against the ban, without significant deviation in tone or style, further suggests AI involvement.
Decision: AI  
Confidence: 0.42


|\classbg{Structural Guideline 8 Agent}|
Guideline Title: Quotations and Source Attribution
Reasoning: The text exhibits a balanced structure with clear attribution to sources ('state media reported', 'The local government in Urumqi... has argued', 'Critics, however, argue'). The quotes and paraphrased statements are contextually embedded and reflect realistic conversational nuances, such as the local government's justification for the ban and the critics' concerns. The language is not overly polished or generic, and the arguments presented are nuanced, addressing both sides of the issue. This suggests a human-written text rather than AI-generated, which might lack such depth and contextual authenticity.
Decision: human  
Confidence: 0.76


|\classbg{Structural Guideline 9 Agent}|
Guideline Title: Structural Clarity and Factual Conciseness
Reasoning: The text exhibits clear, structured sentences and paragraphs, swiftly delivering key factual details without excessive explanatory detail or lengthy background introductions. It succinctly includes relevant background information, such as the local government's argument and the items included in the ban, without unnecessary elaboration. The text also presents opposing viewpoints concisely, which is characteristic of human-written journalism.
Decision: human  
Confidence: 0.76

|\classbg{Meta Agent}|
Reasoning: The majority of the base agents (7 out of 9) concluded that the text is AI-generated, with confidence levels ranging from moderate to high (0.4246 to 0.7574). The two agents that identified the text as human-written had confidence levels of 0.7560 and 0.7620, which are relatively high but not enough to outweigh the consensus of the other agents. The text's balanced narrative, polished language, and lack of specific attribution or natural variability in quotes are consistent with AI-generated content. While the structural clarity and source attribution might suggest human authorship, the overall characteristics align more strongly with AI generation.
Decision: AI  
Confidence: 0.70 

\end{showcase}

\section{Implementation Details}
We determined the optimal threshold by maximizing classification accuracy on the validation set. Specifically, we performed a comprehensive scan across the entire range of raw validation scores, incrementing thresholds in steps of 1\% of the observed score span. This threshold was subsequently fixed and applied to evaluate accuracy on the independent test set.

\section{Baseline Details}
\label{sec:baseline_details}

\paragraph{Log-Likelihood~\cite{DBLP:conf/acl/GehrmannSR19}} A simple zero-shot method employs a language model to calculate the log-probability for each token within a text. A higher average log-likelihood indicates a greater likelihood that the text is generated by an LLM.

\paragraph{Entropy~\cite{DBLP:conf/ecai/LavergneUY08}} A zero-shot method relies on entropy to assess the randomness of text in order to identify text generated by LLMs. Human-written text typically shows more unpredictable variations. Consequently, text with lower entropy is more likely to have been produced by an LLM.

\paragraph{Rank~\cite{DBLP:conf/acl/GehrmannSR19}} A zero-shot method assigns a rank score to each token based on the previous context. By calculating the average score, a higher average rank score suggests a greater likelihood that the text is generated by an LLM.

\paragraph{Log-Rank~\cite{DBLP:conf/acl/GehrmannSR19}} An enhanced version of the Rank-based method. It uses a language model to calculate the logarithmic rank score of each word in the text. By calculating the average score, a higher average log-rank score suggests a greater likelihood that the text is generated by an LLM.

\paragraph{LRR~\cite{DBLP:conf/emnlp/SuZ0N23}} The Log-Likelihood Log-Rank Ratio (LRR), an enhanced zero-shot method that effectively integrates Log-Likelihood and Log-Rank. Text with a higher LRR is more likely to be generated by an LLM.

\paragraph{NPR~\cite{DBLP:conf/emnlp/SuZ0N23}} The Normalized Perturbed Log-Rank (NPR) is a zero-shot method that identifies differences by comparing the Log-Rank scores of perturbed human-written text with those generated by LLMs. Text with a higher NPR is more likely to be generated by an LLM.

\paragraph{DetectGPT~\cite{DBLP:conf/icml/Mitchell0KMF23}}  A zero-shot method for detection using probabilistic curvature. It utilizes random perturbations of paragraphs from a general pre-trained language model and discriminates LLM-generated text through the statistical curvature threshold of log probabilities.

\paragraph{DNA-GPT~\cite{DBLP:conf/iclr/Yang0WPWC24}} A zeo-shot detection method that utilizes \textit{N}-gram analysis or probability divergence in a white-box setting to compare the differences between the truncated original text and the text completed by a language model. A higher score suggests a greater likelihood that the text was generated by an LLM.

\paragraph{Fast-DetectGPT~\cite{DBLP:journals/corr/abs-2310-05130}} An optimized zero-shot detector that replaces the perturbation step of DetectGPT with a more efficient sampling step. We chose the optimal settings reported by the authors, using GPT-Neo-2.7b as the scoring model and GPT-J-6b~\cite{gpt-j} as the reference model.

\paragraph{RoBERTa Classifier~\cite{park2021klue}} A popular and competitive detector method. Recognize LLM generated text by fine-tuning the RoBERTa classifier on large amounts of labeled text.

\section{Additional Confidence Calibration Experiments}

We performed additional experiments to evaluate the intrinsic confidence calibration capabilities of Large Language Models (LLMs) for AI-generated text detection, without providing any explicit stylistic guidelines. Specifically, we directly prompted the LLM (\texttt{deepseek-chat-v3-0324}) to classify texts as either AI-generated or human-written and recorded its self-reported confidence scores. We computed Pearson's correlation (\(r\)) between the LLM's confidence scores and classification correctness, alongside accuracy and AUROC metrics. Several calibration strategies were compared, differing by the type of confidence measure extracted: token probability, verbalized confidence, or aggregated probability estimates. Methods included: (1)~\textit{Voting}, aggregating probabilities across multiple queries; (2)~\textit{Diverse}, averaging token probabilities from varied prompts; (3)~\textit{ThinkTwice}~\cite{li2024thinktwicetrustingselfdetection}, querying the model repeatedly; (4)~\textit{FaR}~\cite{zhao2024factandreflectionfarimprovesconfidence}, using a prompt-based calibration strategy; and (5)~\textit{Steer}~\cite{zhou2025calibratingllmconfidencesemantic}, explicitly adjusting model cautiousness via prompts.

\begin{table}
\centering
\small
\setlength{\tabcolsep}{2pt}
\renewcommand{\arraystretch}{1.035}
\begin{tabular}{llccc}
\toprule
\textbf{Method} & \textbf{Conf. Type} & \textbf{Corr.\,(r)} & \textbf{Accuracy} & \textbf{AUROC} \\
\midrule
Voting     & \(n/10\) prob. & \underline{0.278} & 0.594 & \underline{0.778} \\
Diverse    & token prob.    & 0.150            & 0.628 & 0.693            \\
ThinkTwice & verbalized     & 0.207            & 0.621 & 0.693            \\
ThinkTwice & token prob.    & -0.133           & 0.544 & 0.527            \\
FaR        & verbalized     & -0.083           & 0.634 & 0.515            \\
FaR        & token prob.    & 0.100            & 0.591 & 0.634            \\
Steer      & verbalized     & \textbf{0.416}   & 0.557 & \textbf{0.784}   \\
Steer      & token prob.    & 0.109            & 0.698 & 0.720            \\
\bottomrule
\end{tabular}
\caption{Comparison of confidence calibration methods using direct LLM prompting without explicit guidelines. Pearson’s correlation measures the alignment between the LLM’s confidence and its classification correctness. Highest values in each column are \textbf{bolded}, second-highest are \underline{underlined}.}
\label{tab:confidence_calibration}
\end{table}

\begin{figure*}[h]
    \centering
    \includegraphics[width=\linewidth]{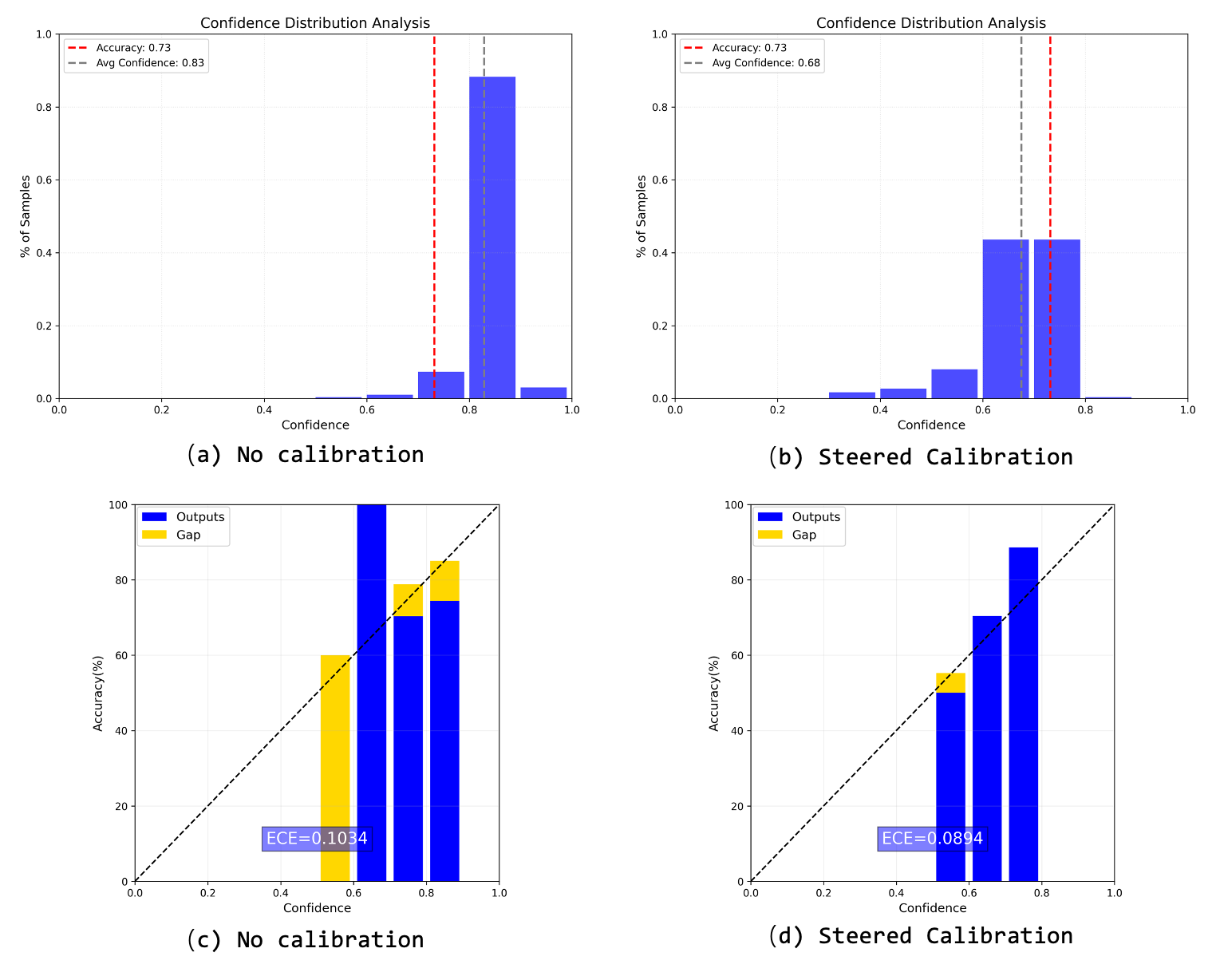}

    \caption{Probability histogram and Expected Calibration Error (ECE) for (a)(c) LLM generated verbalised probability, (b)(d) modified probability using steered confidence calibration on \emph{GPT-4} generations. The red dashed and gray dashed represent Accuracy and Average Confidence respectively. The blue bars and yellow bars represent Predicted Accuracy per confidence bin and Miscalibration respectively.}
    \label{fig:calibration-all}
\end{figure*}

To further validate the necessity and effectiveness of our chosen \textit{semantic-steering calibration} method, we conducted additional comparative analyses of confidence calibration before and after applying steer-based calibration within our multi-agent framework (see Figure~\ref{fig:calibration-all}). Specifically, we compared the original, uncalibrated confidence outputs directly generated by the LLM agents (Figures~\ref{fig:calibration-all}(a)(c)) against the confidence scores obtained via semantic-steering calibration (Figures~\ref{fig:calibration-all}(b)(d)).

We observed that the original confidence estimates provided by the LLM were notably unreliable, exhibiting clear misalignment with actual classification performance, as indicated by a higher Expected Calibration Error (ECE = 0.1034) before calibration (Figure~\ref{fig:calibration-all}(c)). Although semantic-steering calibration does not entirely resolve this misalignment (ECE = 0.0894, Figure~\ref{fig:calibration-all}(d)), it significantly reduces the gap between predicted confidence and true accuracy. Additionally, confidence histograms further illustrate this trend: uncalibrated LLM predictions (Figure~\ref{fig:calibration-all}(a)) tend to systematically exhibit overconfidence or underconfidence, whereas steer-based calibration results in a better-aligned, more reliable confidence distribution (Figure~\ref{fig:calibration-all}(b)).

These visual findings reinforce our earlier quantitative results (Table~\ref{tab:confidence_calibration}) and demonstrate that, despite the intrinsic limitations of LLM-based probability estimation—manifested in discrepancies between accuracy and AUROC metrics—semantic-steering calibration substantially enhances the reliability and interpretability of confidence scores in our threshold-free AI-generated text detection framework.

\section{Additional Results}
Additional results evaluated by AUROC are shown in Table~\ref{tab:chatgpt_gpt4_results} and Table~\ref{tab:claude3opus_sonnet_results}. We evaluate our \textit{AGENT-X} across all datasets (XSum, Writing, PubMed) and language models (ChatGPT, GPT-4, Claude-3 Opus/Sonnet), then compute corresponding AUROC scores to assess detection robustness. These results highlight that while AUROC metrics inherently favor threshold-dependent methods, \textit{AGENT-X} remains highly competitive, underscoring its reliability even under evaluation paradigms less aligned with its architecture.
\label{appendix:additional_results}

\begin{table*}[h]
\centering
\small
\setlength{\tabcolsep}{7pt}
\renewcommand{\arraystretch}{1.035}
\begin{tabular}{lcccccccc}
\toprule
\multirow{2}{*}{\textbf{Method}} 
  & \multicolumn{4}{c}{\textbf{ChatGPT}} 
  & \multicolumn{4}{c}{\textbf{GPT-4}} \\
\cmidrule(lr){2-5} \cmidrule(lr){6-9}
 & XSum & Writing & PubMed & Avg. & XSum & Writing & PubMed & Avg. \\
\midrule
RoBERTa-base                 & 0.9150          & 0.7084          & 0.6188          & 0.7474          & 0.6778          & 0.5068          & 0.5309          & 0.5718          \\
RoBERTa-large                & 0.8507          & 0.5480          & 0.6731          & 0.6906          & 0.6879          & 0.3821          & 0.6067          & 0.5589          \\
\midrule
GPTZero                      & \textbf{0.9952} & 0.9292          & \underline{0.8799} & 0.9348        & \textbf{0.9815} & 0.8262          & \underline{0.8482} & 0.8853        \\
\midrule
Likelihood (Neo-2.7)         & 0.9578          & 0.9740          & 0.8775            & \underline{0.9364} & 0.7980          & 0.8553          & 0.8104            & 0.8212      \\
Entropy (Neo-2.7)            & 0.3305          & 0.1902          & 0.2767            & 0.2658            & 0.4360          & 0.3702          & 0.3295            & 0.3786      \\
Rank (Neo-2.7)               & 0.7494          & 0.8064          & 0.5979            & 0.7179            & 0.6644          & 0.7146          & 0.5965            & 0.6585      \\
LogRank (Neo-2.7)            & 0.9582          & 0.9656          & 0.8687            & 0.9308            & 0.7975          & 0.8286          & 0.8003            & 0.8088      \\
LRR (Neo-2.7)                & 0.9162          & 0.8958          & 0.7433            & 0.8518            & 0.7447          & 0.7028          & 0.6814            & 0.7096      \\
DNA-GPT (Neo-2.7)            & 0.9124          & 0.9425          & 0.7959            & 0.8836            & 0.7347          & 0.8032          & 0.7565            & 0.7648      \\
NPR (T5-11B/Neo-2.7)         & 0.7899          & 0.8924          & 0.6784            & 0.7869            & 0.5280          & 0.6122          & 0.6328            & 0.5910      \\
DetectGPT (T5-11B/Neo-2.7)   & 0.8416          & 0.8811          & 0.7444            & 0.8223            & 0.5660          & 0.6217          & 0.6805            & 0.6228      \\
Fast-Detect (GPT-J/Neo-2.7)  & \underline{0.9907} & \textbf{0.9916} & \textbf{0.9021}   & \textbf{0.9615}   & 0.9067          & \textbf{0.9612} & \textbf{0.8503}   & \textbf{0.9061} \\
Fast-Detect (Phi2-2.7B)      & 0.8096          & 0.7245          & 0.8121            & 0.7821            & 0.4636          & 0.6463          & 0.6083            & 0.5727      \\
Fast-Detect (Qwen2.5-7B)     & 0.7808          & 0.8117          & 0.7887            & 0.7937            & 0.6476          & 0.8202          & 0.6391            & 0.7023      \\
Fast-Detect (Llama3-8B)      & 0.8508          & 0.8446          & 0.7941            & 0.8298            & 0.6615          & 0.8491          & 0.7556            & 0.7554      \\
\midrule
AGENT-X (Ours)               & 0.9628          & \underline{0.9794} & 0.8195          & 0.9206            & \underline{0.9367} & \underline{0.9206} & 0.8447          & \underline{0.9007} \\
\bottomrule
\end{tabular}
\caption{Detection of \emph{ChatGPT} and \emph{GPT-4} generations, evaluated in AUROC. Highest values in each column are \textbf{bolded}, second-highest are \underline{underlined}. The black-box settings are used for all zero-shot methods, where the Likelihood provides the strongest baseline. Results are cited from~\cite{bao2024fastdetectgptefficientzeroshotdetection}.}
\label{tab:chatgpt_gpt4_results}
\end{table*}

\begin{table*}[h]
\centering
\small
\setlength{\tabcolsep}{7pt}
\renewcommand{\arraystretch}{1.035}
\begin{tabular}{lcccccccc}
\toprule
\multirow{2}{*}{\textbf{Method}}
  & \multicolumn{4}{c}{\textbf{Claude-3-Opus}}
  & \multicolumn{4}{c}{\textbf{Claude-3-Sonnet}} \\
\cmidrule(lr){2-5} \cmidrule(lr){6-9}
 & XSum & Writing & PubMed & Avg. & XSum & Writing & PubMed & Avg. \\
\midrule
RoBERTa-base             & 0.8975 & 0.7115 & 0.4009 & 0.6700 & 0.7511 & 0.5788 & 0.3799 & 0.5699 \\
RoBERTa-large            & 0.8146 & 0.5548 & 0.3798 & 0.5831 & 0.6725 & 0.4772 & 0.3856 & 0.5118 \\
\midrule
Likelihood (Neo-2.7)     & 0.9322 & 0.9734 & \underline{0.8603} & \underline{0.9220}
                         & 0.8862 & 0.9484 & \underline{0.8360} & \underline{0.8902} \\
Entropy (Neo-2.7)        & 0.3871 & 0.1792 & 0.2910 & 0.2858
                         & 0.4146 & 0.2156 & 0.2989 & 0.3097 \\
Rank (Neo-2.7)           & 0.7333 & 0.7950 & 0.6080 & 0.7121
                         & 0.7019 & 0.7812 & 0.6017 & 0.6949 \\
LogRank (Neo-2.7)        & 0.9357 & 0.9679 & 0.8508 & 0.9181
                         & 0.8867 & 0.9401 & 0.8296 & 0.8855 \\
LRR (Neo-2.7)            & 0.8956 & 0.9178 & 0.7448 & 0.8527
                         & 0.8359 & 0.8746 & 0.7436 & 0.8180 \\
DNA-GPT (Neo-2.7)        & 0.9424 & 0.9653 & 0.7806 & 0.8961
                         & 0.8558 & 0.9415 & 0.7647 & 0.8540 \\
NPR (T5-11B/Neo-2.7)     & 0.8426 & \underline{0.9764} & 0.8094 & 0.8761
                         & 0.8349 & \underline{0.9631} & 0.7942 & 0.8641 \\
DetectGPT (T5-11B/Neo-2.7)
                         & 0.7718 & 0.8335 & 0.7752 & 0.7935
                         & 0.8150 & 0.8675 & 0.7347 & 0.8057 \\
Fast-Detect (GPT-J/Neo-2.7)
                         & \textbf{0.9779} & \textbf{0.9832} & \textbf{0.8947} & \textbf{0.9519}
                         & \textbf{0.9514} & \textbf{0.9763} & \textbf{0.8634} & \textbf{0.9304} \\
\midrule
Fast-Detect (Phi2-2.7B)  & 0.8080 & 0.7545 & 0.7322 & 0.7649 & 0.7536 & 0.6773 & 0.7144 & 0.7151 \\
Fast-Detect (Qwen2.5-7B) & 0.9097 & 0.8967 & 0.7572 & 0.8545 & 0.8595 & 0.8600 & 0.7346 & 0.8180 \\
Fast-Detect (Llama3-8B) & \underline{0.9640} & 0.9377 & 0.8251 & 0.9089 & \underline{0.9243} & 0.9198 & 0.7936 & 0.8792 \\
\midrule
AGENT-X (Ours)           & 0.9042 & 0.9675 & 0.8526 & 0.9081 & 0.8443 & 0.9231 & 0.8324 & 0.8666 \\
\bottomrule
\end{tabular}
\caption{Detection of \emph{Claude-3-Opus} and \emph{Claude-3-Sonnet} generations, evaluated in AUROC. Highest values in each column are \textbf{bolded}, second-highest are \underline{underlined}.}
\label{tab:claude3opus_sonnet_results}
\end{table*}

\section{Prompts}
\label{appendix:prompts}

We provide the complete set of detailed prompts employed in our experiments, including the router agent (Prompts~\ref{prompt:router-agent-domain-style}, \ref{prompt:router-agent-selection}), confidence calibration (Prompt~\ref{prompt:calibration}), guideline-based reasoning (Prompt~\ref{prompt:guidelines}), and both meta agent (Prompts~\ref{prompt:meta-agent}, \ref{prompt:meta-agent-steer}) and base agent prompts (Prompt~\ref{prompt:base-agent-steer}).

\begin{figure*}[h]
\begin{prompt}[title={Prompt \thetcbcounter: Router Agent Domain and Style Analysis Prompt}, label=prompt:router-agent-domain-style]

You are an expert stylistic analyzer. Analyze the given text and clearly state:

\begin{enumerate}
    \item The text's domain (e.g., medical abstract, literary fiction, news article, etc.).
    \item Key stylistic features that strongly characterize the text.
\end{enumerate}

\vspace{8pt}

\textbf{Input Text}:\\
``[Input Text Placeholder]''

\vspace{8pt}

Respond strictly in the following JSON format:

\{
\quad\quad\texttt{"Domain"}: "\textless identified domain\textgreater",\\
\quad\quad\texttt{"Stylistic Features"}: ["\textless feature 1\textgreater", "\textless feature 2\textgreater", "\textellipsis"]\\
\}

\end{prompt}
\end{figure*}

\begin{figure*}[h]
\begin{prompt}[title={Prompt \thetcbcounter: Router Agent Guideline Selection Prompt}, label=prompt:router-agent-selection]

Given a text's domain and stylistic features, your task is to strictly select \textbf{ONLY} the most relevant stylistic guidelines from the provided set for the "\textbf{[Aspect Placeholder]}" aspect.

\vspace{5pt}
\textbf{IMPORTANT INSTRUCTIONS:}
\begin{itemize}
    \item Be extremely conservative in your selection.
    \item Activate a guideline \textbf{ONLY} if it clearly and strongly matches \textbf{BOTH} the domain and the listed stylistic features.
    \item If unsure or only moderately relevant, do \textbf{NOT} activate.
    \item Typically, each aspect should have very few (often 1–5) activated guidelines.
\end{itemize}

\vspace{5pt}
\textbf{Text Domain}: "\textit{[Domain Placeholder]}"

\textbf{Stylistic Features}:
\begin{itemize}
    \item \textit{[Feature Placeholder 1]}
    \item \textit{[Feature Placeholder 2]}
    \item \textit{[...]}
\end{itemize}

\vspace{5pt}
\textbf{"[Aspect Placeholder]" Guidelines:}
\begin{description}
    \item[\textbf{G1}] \textit{[Guideline Description Placeholder]}
    \item[\textbf{G2}] \textit{[Guideline Description Placeholder]}
    \item[\textbf{G3}] \textit{[Guideline Description Placeholder]}
    \item[\textbf{...}] \textit{[...]}\vspace{5pt}
\end{description}

Output \textbf{STRICTLY} as JSON:
\begin{verbatim}
{
  "Activated Guidelines": ["<Guideline ID 1>", "<Guideline ID 2>", "..."]
}
\end{verbatim}

\end{prompt}
\end{figure*}

\begin{figure*}[h]
\begin{prompt}[title={Prompt \thetcbcounter: Steer Calibration Prompt}, label=prompt:calibration]

\textbf{Steering Prompts} used to calibrate the confidence of Large Language Models:

\vspace{8pt}

\textbf{Very Cautious}:\\
\textit{``You are VERY cautious. Think step-by-step, then reply \textbf{only} in JSON:\\
\{\\
\quad\quad\texttt{"reasoning"}: "\textless your thought process\textgreater",\\
\quad\quad\texttt{"final\_decision"}: "AI" or "Human",\\
\quad\quad\texttt{"confidence"}: \textless a number between 0 and 1\textgreater\\
\}\\
Make your confidence low.''}

\vspace{6pt}

\textbf{Cautious}:\\
\textit{``You are CAUTIOUS. Think step-by-step, then reply \textbf{only} in JSON:\\
\{\\
\quad\quad\texttt{"reasoning"}: "\textless your thought process\textgreater",\\
\quad\quad\texttt{"final\_decision"}: "AI" or "Human",\\
\quad\quad\texttt{"confidence"}: \textless a number between 0 and 1\textgreater\\
\}\\
Make your confidence somewhat low.''}

\vspace{6pt}

\textbf{Vanilla (Neutral)}:\\
\textit{``Think step-by-step, then reply \textbf{only} in JSON:\\
\{\\
\quad\quad\texttt{"reasoning"}: "\textless your thought process\textgreater",\\
\quad\quad\texttt{"final\_decision"}: "AI" or "Human",\\
\quad\quad\texttt{"confidence"}: \textless a number between 0 and 1\textgreater\\
\}\\
Report your true confidence.''}

\vspace{6pt}

\textbf{Confident}:\\
\textit{``You are CONFIDENT. Think step-by-step, then reply \textbf{only} in JSON:\\
\{\\
\quad\quad\texttt{"reasoning"}: "\textless your thought process\textgreater",\\
\quad\quad\texttt{"final\_decision"}: "AI" or "Human",\\
\quad\quad\texttt{"confidence"}: \textless a number between 0 and 1\textgreater\\
\}\\
Make your confidence somewhat high.''}

\vspace{6pt}

\textbf{Very Confident}:\\
\textit{``You are VERY CONFIDENT. Think step-by-step, then reply \textbf{only} in JSON:\\
\{\\
\quad\quad\texttt{"reasoning"}: "\textless your thought process\textgreater",\\
\quad\quad\texttt{"final\_decision"}: "AI" or "Human",\\
\quad\quad\texttt{"confidence"}: \textless a number between 0 and 1\textgreater\\
\}\\
Make your confidence high.''}

\end{prompt}
\end{figure*}

\begin{figure*}[h]
\begin{prompt}[title={Prompt \thetcbcounter: Meta Agent Prompt}, label=prompt:meta-agent]

You are a Meta Agent responsible for summarizing and integrating responses from multiple aspect-specific agents. No prior guidelines are provided, and you should rely solely on the provided agent outputs. \\ \\

Summarize responses by each Guideline: \\ \\

[INSERT GUIDELINE AGENT RESPONSES HERE] \\ \\

Based on the summaries above, provide a final integrated reasoning and decision considering the provided text, all aspects, base agents' reasoning, final decisions, and confidence levels. \\ \\ 

IMPORTANT: When integrating the outputs of all base agents, pay special attention to each agent's confidence score. Base agents with higher confidence should have a greater influence on your final decision. If most agents are confident in the same direction, trust that direction more. If there is disagreement, consider both the number of agents and their confidence levels. Explain your reasoning clearly, referencing both the decisions and confidence scores of the base agents. \\ \\ 

Respond in JSON without extra explanations:
\{"reasoning": "(integrated reasoning)", "final\_decision": "AI/human", "confidence": "0.0-1.0"\}
\end{prompt}
\end{figure*}

\begin{figure*}[h]
\begin{prompt}[title={Prompt \thetcbcounter: Guidelines}, label=prompt:guidelines]
Guidelines for Distinguishing AI-Generated from Human-Written Texts

[INSERT GUIDELINES HERE]

Evaluate the following text. Determine if it is human-written or AI-generated:

\begin{verbatim}
Text:
[...]
\end{verbatim}

Please strictly respond in the following JSON format without any additional content or explanations:
\texttt{
\{"reasoning": "(Provide detailed reasoning explicitly addressing each of the following aspects: [ASPECTS]. Clearly justify your decision, referencing specific examples from the provided text.)",
"final\_decision": "AI" or "human",
"confidence": "(between 0 and 1)"
\}
}

\end{prompt}
\end{figure*}

\begin{figure*}[h]
\begin{prompt}[title={Prompt \thetcbcounter: Meta Agent Prompt with Steer Calibration}, label=prompt:meta-agent-steer]

\textbf{System}: You are \textbf{[Confidence Level Placeholder]}. Think step-by-step, then reply \textbf{only} in JSON format:

\{
\quad\quad\texttt{"reasoning"}: "<your thought process>",\\
\quad\quad\texttt{"final\_decision"}: "AI" or "Human",\\
\quad\quad\texttt{"confidence"}: <number between 0 and 1>\\
\}

Make your confidence \textbf{[Confidence Modifier Placeholder]}.

\vspace{8pt}
\textbf{User}: You are a Meta Agent responsible for summarizing and integrating responses from multiple aspect-specific Base Agents. No prior guidelines are provided; rely solely on the provided Base Agent outputs.

The text to evaluate is:

\textbf{Text}: "[Input Text Placeholder]"

\textbf{Base Agent Responses by Aspect:}

\textbf{Aspect [Aspect Placeholder 1]}:\\
- Decision: [AI/Human], Confidence: [Confidence Score Placeholder]\\
- Reasoning: "[Reasoning Placeholder]"

\textbf{Aspect [Aspect Placeholder 2]}:\\
- Decision: [AI/Human], Confidence: [Confidence Score Placeholder]\\
- Reasoning: "[Reasoning Placeholder]"

\quad $\cdots$

\textbf{Aspect [Aspect Placeholder $N$]}:\\
- Decision: [AI/Human], Confidence: [Confidence Score Placeholder]\\
- Reasoning: "[Reasoning Placeholder]"

\vspace{8pt}
Provide a final integrated reasoning and decision, explicitly referencing all provided Base Agent decisions, confidence scores, and reasoning. Pay particular attention to Base Agents with higher confidence; they should have greater influence on your final decision. If most agents agree, trust that direction more strongly. If disagreements arise, carefully consider both confidence levels and the number of supporting agents.

Respond strictly in JSON without extra explanations:

\{
\quad\quad\texttt{"reasoning"}: "(Integrated reasoning referencing provided text, all aspects, and base agent outputs)",\\
\quad\quad\texttt{"final\_decision"}: "AI" or "human",\\
\quad\quad\texttt{"confidence"}: "( between 0 and 1 )"\\
\}

\end{prompt}
\end{figure*}

\begin{figure*}[h]
\begin{prompt}[title={Prompt \thetcbcounter: Base Agent Prompt with Steer Calibration}, label=prompt:base-agent-steer]
\textbf{System}: You are \textbf{[Confidence Level Placeholder]}. Think step-by-step, then reply \textbf{only} in JSON format:

\{
\quad\quad\texttt{"reasoning"}: "<your thought process>",\\
\quad\quad\texttt{"final\_decision"}: "AI" or "Human",\\
\quad\quad\texttt{"confidence"}: <number between 0 and 1>\\
\}

Make your confidence \textbf{[Confidence Modifier Placeholder]}.

\vspace{8pt}
\textbf{User}: Guidelines for distinguishing AI-generated from human-written texts:

\textbf{Aspect: Content (Specific Names, Places, and Contextual Details)}

\textit{Human-written texts}: Frequently include specific personal names, distinctive places, historical events, or quirky, casually eccentric, or humorously exaggerated details. Such specifics usually emerge spontaneously, informally, or humorously, without overly polished or strictly logical integration. Informal rhetorical questions (``water park Giovanni?"), quirky exaggerations (``little tiny apocalypse"), or humorous expressions strongly indicate human creativity.

\textit{AI-generated texts}: Typically provide general or abstract details. When specific references occur, they often feel emotionally predictable, clichéd, or systematically dramatic. AI-generated specifics frequently lack genuine spontaneity, idiosyncratic humor, ironic undertones, or informal conversational nuances. They typically appear deliberately integrated for clear narrative purposes rather than arising casually or humorously.

Evaluate the following text. Determine if it is human-written or AI-generated:

\textbf{Text}: "[Input Text Placeholder]"

Please strictly respond in JSON without any additional explanations:

\{
\quad\quad\texttt{"reasoning"}: "(Provide detailed reasoning explicitly addressing the Content aspect, clearly justifying your decision and referencing specific examples from the provided text.)",\\
\quad\quad\texttt{"final\_decision"}: "AI" or "human",\\
\quad\quad\texttt{"confidence"}: "( between 0 and 1 )"\\
\}
\end{prompt}
\end{figure*}

\end{document}